# Machining Cycle Time Prediction: Data-driven Modelling of Machine Tool Feedrate Behavior with Neural Networks

Chao Sun[1], Javier Dominguez-Caballero[1], Rob Ward[1], Sabino Ayvar-Soberanis[1], David Curtis[1]

[1] *Advanced Manufacturing Research Centre, University of Sheffield, Rotherham, S60 5TZ, UK*

## Abstract

Accurate prediction of machining cycle times is important in the manufacturing industry. Usually, Computer-Aided Manufacturing (CAM) software estimates the machining times using the commanded feedrate from the toolpath file using basic kinematic settings. Typically, the methods do not account for toolpath geometry or toolpath tolerance and therefore underestimate the machining cycle times considerably. Removing the need for machine-specific knowledge, this paper presents a data-driven feedrate and machining cycle time prediction method by building a neural network model for each machine tool axis. In this study, datasets composed of the commanded feedrate, nominal acceleration, toolpath geometry and the measured feedrate were used to train a neural network model. Validation trials using a representative industrial thin-wall structure component on a commercial machining centre showed that this method estimated the machining time with more than 90% accuracy. This method showed that neural network models have the capability to learn the behavior of a complex machine tool system and predict cycle times. Further integration of the methods will be critical in the implantation of digital twins in Industry 4.0.

## 1. Introduction

In the era of Industry 4.0, the digitisation of the production process is driving innovation, providing new tools for optimising manufacturing processes, supporting business critical decision-making and ultimately adding value to the production process. Traditionally, process modelling-based technologies were classed as either offline or online, whereas now hybrid solutions are offering a more symbiotic relationship where online data can support offline analysis which in turn can optimise online processes. Currently, the technology Zeitgeist is the Digital Twin [1][2] and data-driven artificial intelligence [3]. Digital Twins support the manufacturing process by concurrently running virtual simulations with the actual manufacturing process in real-time using live plant information. It has been utilized to predict maintenance[4], analyse assembly precision[5] etc. The accuracy and fidelity of the Digital Twins are only as accurate as the simulation models driving them [6]. Therefore, the Digital Twin is dependent upon expert knowledge and advanced modelling techniques specific to the process. One method of overcoming this requirement is through the use of data-driven artificial intelligence modelling. The process to be modelled can be thought of as a black box with measurable inputs and outputs [7]. The task of artificial intelligence based modelling is to map the inputs to the outputs and predict the system responses to future inputs without the requirement for advanced process knowledge. Data-driven artificial intelligence modelling has been utilised for tool condition forecasting [8], robot fault diagnosis [9], surface defect recognition [10], and machining parameter planning [11].

Within the machining industry, the ability to predict accurate machining cycle times is vital to ensuring realistic and profitable contracts, supporting production and supply chain planning. Traditionally, predicting accurate machining cycle times has been poor. Within Computer Aided Manufacture (CAM) software, the machining time is usually estimated with the assumption that the machine tool can always attain the commanded feedrate set in the part program and does not account for toolpath geometry. However, in high-speed machining during complex toolpaths, the actual feedrate is usually far lower than the commanded feedrate. One study showed that when the evaluation was only based on the commanded values, the error for machining time prediction exceeded 36% [12]. This is due to the Numerical Controller (NC) interpolator taking into account the kinematic limits of the machine tool feed drives (jerk and maximum feedrates), part geometry and tool centre point (TCP) tolerance during trajectory generation.

Early researchers approached the problem of machining cycle estimation not from a feedrate perspective but from a holistic analysis of the toolpath characteristics. One method predicted machining time using the frequency distribution of cutter location line lengths [13] and another added kinematics to the distribution-based approach whilst recognising that cornering feedrate was an important factor [14]. An approach based on the material removal volume and specified the surface roughness of each feature was proposed [15]. A feature-based method for NC machining time estimation was proposed by using geometry-process information and the machine characteristics, and in the prototype system, an error of less than 5% was reached [16]. Also, machining time was calculated holistically using the machine response time, which is a machine-specific parameter calculated from a prior system testing [17]. More recently, another holistic approach used the upper feedrate limit and experimentally determined sampling time to determine the overall machining cycle time [18].

Approaching the problem through trajectory generation and feedrate prediction, one method is to build virtual machine tool models using analytical methods. In past research, models were proposed for point-to-point interpolation based on jerk limit [19][20]. For a point-to-point (P2P) interpolation, it is assumed that the tool starts from a full-stop at one point and comes to a full-stop at the end of the cutter location line, this can provide accurate predictions of cycle time if the machine is set to P2P motion. However, in modern NC systems, the look-ahead function enables non-stop smooth feed motion. Therefore to model the non-stop interpolation behavior of modern NC systems significant research has been conducted in non-stop interpolation and geometric blending methods. The research spans from circular arcs, cubic [21] and quintic splines [22] through to modern Finite Impulse Response based filtering methods [23] [24]. The modern Finite Impulse Response based filtering methods gave more than 90% accuracy on machining time prediction.

Another possible method of approaching machining cycle time estimation is by using data-driven methods. Compared to the analytical methods, the main benefit of using data-driven methods is it does not require the user to have advanced control knowledge to do system identification. The machine tool can be regarded as a black box for which system identification methods can be used to find its transfer function. Feedforward neural networks are capable of approximating a function and its derivatives to be arbitrarily accurate[25]. Therefore, Neural networks were used for nonlinear dynamic system modeling [26]. As shown in Figure 1, compared to holistic and analytical methods which usually require experienced engineers to do system behavior identification and build the model, data-driven models take the system as a black box and do not need much experience from the user. The disadvantage of a data-driven model is that it requires a reasonable amount of data. However, nowadays, many industry solutions have provided the possibility of having access to a huge amount of data from machine tools [27][28].

There have been some Artificial Intelligence based approaches to cycle time estimation albeit with limited utility and verification. For example, deep neural networks have been explored using the workpiece shape and the complexity of the toolpath etc. [29]. The method was validated on three different curves but validated predictions to more general toolpaths have yet to be published. Convolutional neural networks have been applied to images of simple curved surfaces to predict cycle time for simple workpieces [30]. Both of the methods were using holistic analysis of the workpiece and the toolpath. There exists a gap in literature developing data-driven artificial intelligence methods to model the machine tool feedrate behavior and then to predict the machining cycle time.

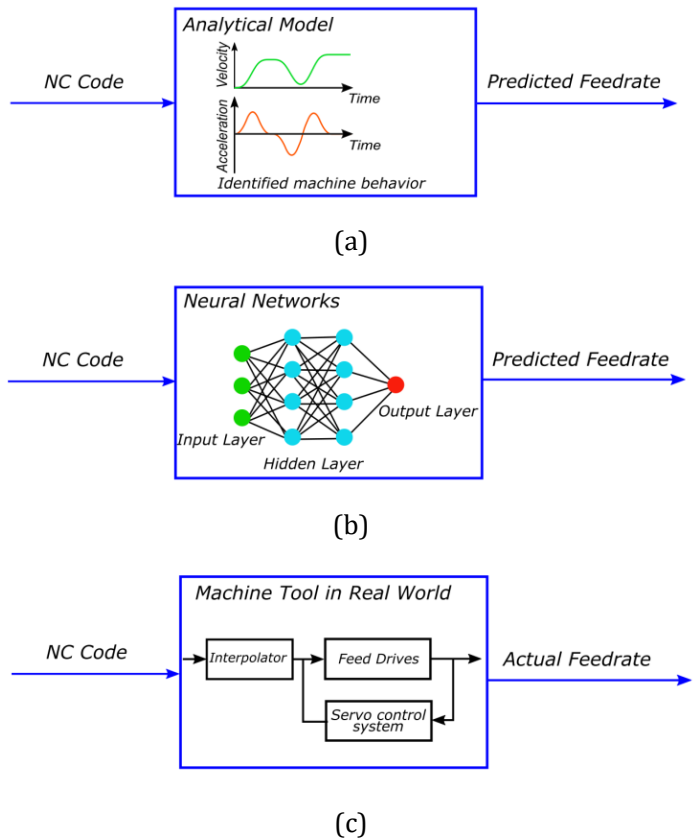

(a)

(b)

(c)

Figure 1. Comparison of models (a) Analytical model (b) Neural network model (c) Machine tool real-world behavior.

This paper proposes a data-driven method to predict the feedrate and machining cycle time by using neural networks to learn the feedrate behavior of the machine tool. As shown in Figure 2, during the machining process the actual feedrate was measured to generate datasets to train the neural network model. The neural network can be used to predict the machining time and could help to optimise the machining process.

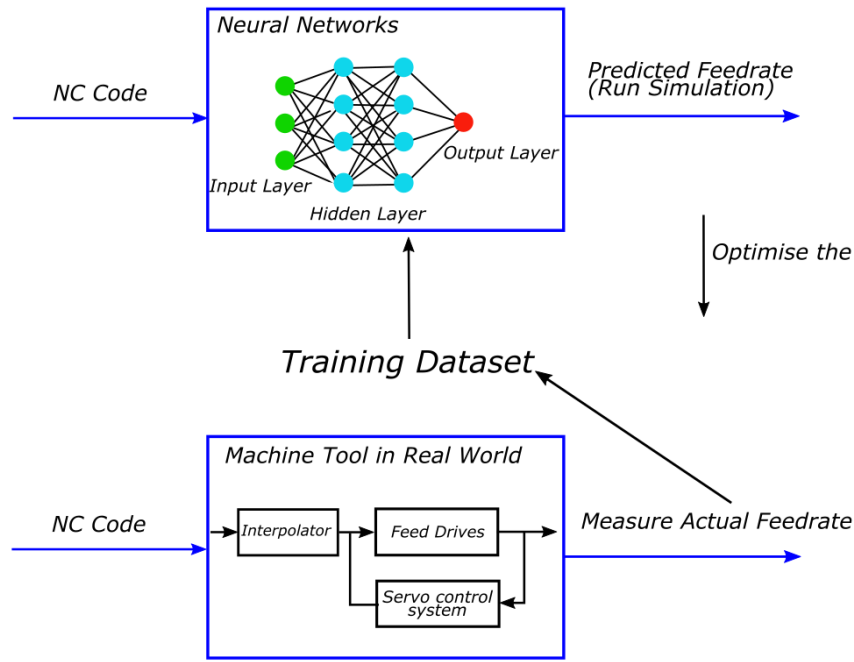

Figure 2. Interactions between the physical part and digital model.

The rest of this article is organized as follows. Section 2 introduces the procedure of building neural networks for feedrate prediction, including the architecture of neural networks, data preprocessing, and training and prediction. Section 3 provides the experimental validation for feedrate and machining cycle time prediction.

## 2. Feedrate and Cycle Time Prediction with Neural Networks

This section introduces the details of building and training a neural network for feedrate and cycle time prediction. In Section 2.1, the architecture of the neural network was discussed. Section 2.2 covered input features and data preprocessing. Section 2.3 introduced the model training and result prediction.

### 2.1 The Architecture of Neural Networks

The architecture of neural networks for feedrate prediction is shown in Figure 3. Each axis of a 3-axis machine tool was modelled by a neural network because the dynamic behavior of each axis can be different. Features, also known as the input of neural networks, like commanded feedrate and nominal acceleration for each axis were extracted from NC code. The details of feature extraction are discussed in Section 2.2. The features are fed into their corresponding neural network to predict the feedrate of each axis. Finally, the resultant feedrate can be calculated with feedrate values from each axis.

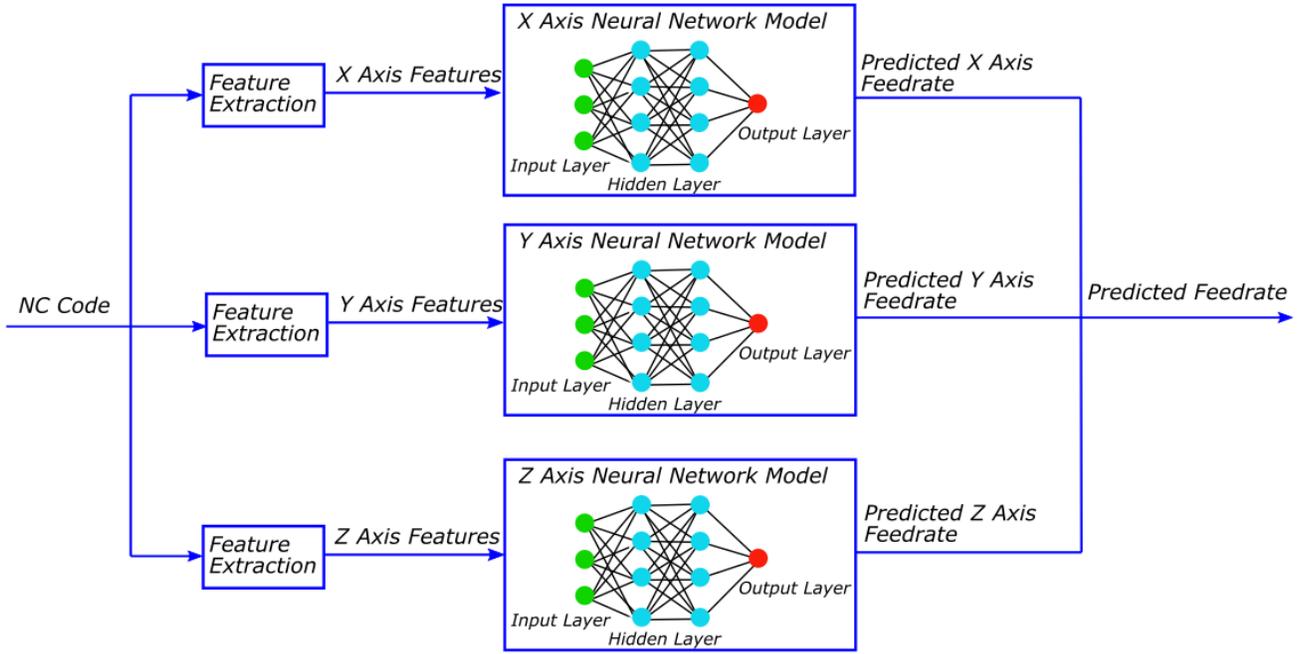

Figure 3. The architecture of neural networks for feedrate prediction.

For each axis, the architecture of a feedforward neural network with $k$ hidden layers is shown in Figure 4. The output of the $s^{th}$ neurons of the first hidden layer can be calculated by equation (1) where $tansig$ is a hyperbolic tangent sigmoid transfer function that introduces non-linearity to the neural network [31]; $p_i$ is the input values to the network; $\omega_{k,s,i}$ is the weight parameters and $b_{k,s}$ is the bias parameters. It should be noted that the number of neurons in each hidden layer can be different. For example, there are $s$ neurons in the first hidden layer and $t$ neurons in the last hidden layer as shown in Figure 4. If there are $m$ inputs coming from the last layer, the output of the $t^{th}$ neurons of the $k^{th}$ hidden layer can be calculated by equation (2). If there are $t$ inputs coming from the last layer, the output of the $r^{th}$ neurons of the output layer can be calculated by equation (3), where $linear$ is the linear transfer fuction [31]. The final output of the neural network can be calculated by equation (4).

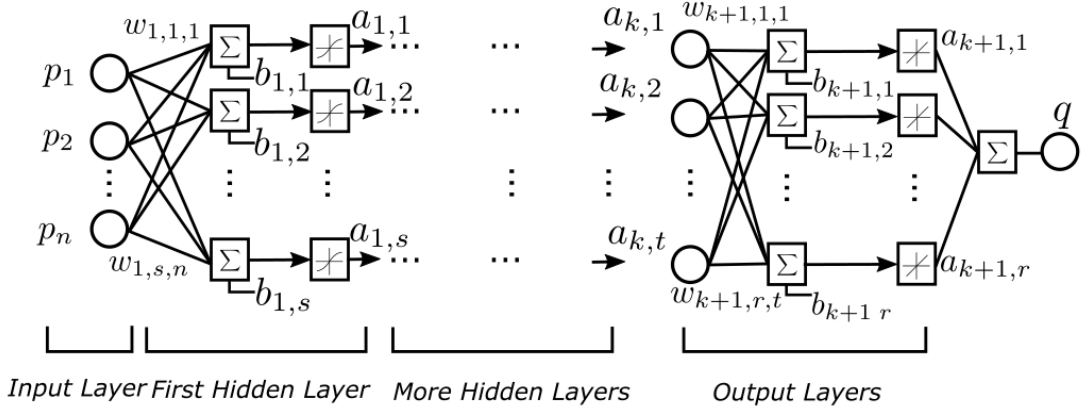

Figure 4. The architecture of the neural network for each axis.

$$a_{k,s} = tansig\left[\sum_{i=1}^{n}(p_i \cdot \omega_{k,s,i}) + b_{k,s}\right] \tag{1}$$

$$a_{k,t} = tansig\left[\sum_{i=1}^{m}(p_i \cdot \omega_{k,t,i}) + b_{k,t}\right] \tag{2}$$

$$a_{k+1,r} = linear\left[\sum_{i=1}^{t}(p_i \cdot \omega_{k+1,r,i}) + b_{k+1,r}\right] \tag{3}$$

$$q = \sum_{i=1}^{r} a_{k+1,i} \tag{4}$$

## *2.2 Input Features and data preprocessing*

Feature extraction is to acquire as much information related to the output actual feedrate as possible from the NC code. Usually, the larger the number of features are, the better the accuracy of prediction that can be achieved. However, the increased number of features will increase the complexity of the neural network and therefore, will require more data for model training. In this paper, the features for each axis of the machine tool are listed in Table 1.

For the $n^{th}$ cutter location, the nominal feedrate and nominal acceleration for each axis, and the angle between line segments at the $(n-1)^{th}$, the $n^{th}$ and the $(n+1)^{th}$ cutter locations are extracted as features to predict the actual feedrate at $n^{th}$ cutter location. The nominal machining time $t_n$ between the $n^{th}$ and the $(n-1)^{th}$ cutter location can be calculated by equation (5) in which $X_n$, $Y_n$, $Z_n$ are the coordinates of the $n^{th}$ cutter location; $X_{n-1}$, $Y_{n-1}$, $Z_{n-1}$ are the coordinates of the $n-1^{th}$ cutter location; $f_n$ is the commanded feedrate for the $n^{th}$ cutter location in NC code. Then the nominal feedrate at the $n^{th}$ cutter location for each axis can be calculated by equation (6) and the nominal acceleration at the $n^{th}$ cutter location for each axis can be calculated by equation (7). The angle at the $n^{th}$ cutter location (Figure 5) can be calculated by equation (8) in which $CL_n$ is the $n^{th}$ cutter location.

$$t_n = \frac{\sqrt{(X_n - X_{n-1})^2 + (Y_n - Y_{n-1})^2 + (Z_n - Z_{n-1})^2}}{f_n} \quad (5)$$

$$\begin{cases} f_{x,n} = \dfrac{(X_n - X_{n-1})}{t_n} \\ f_{y,n} = \dfrac{(Y_n - Y_{n-1})}{t_n} \\ f_{z,n} = \dfrac{(Z_n - Z_{n-1})}{t_n} \end{cases} \quad (6)$$

$$\begin{cases} a_{x,n} = \dfrac{(f_{x,n} - f_{x,n-1})}{t_n} \\ a_{y,n} = \dfrac{(f_{y,n} - f_{y,n-1})}{t_n} \\ a_{z,n} = \dfrac{(f_{z,n} - f_{z,n-1})}{t_n} \end{cases} \quad (7)$$

$$\theta_n = \frac{(CL_{n+1} - CL_n) \cdot (CL_{n-1} - CL_n)}{|CL_{n+1} - CL_n| \cdot |CL_{n-1} - CL_n|} \quad (8)$$

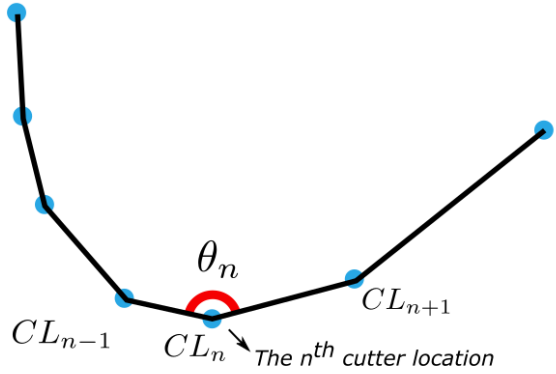

Figure 5. The angle between line segments.

To generate the output (also known as target in neural networks) of the training dataset for the neural network model, the measured actual feedrate at the cutter location is also needed. However, as shown in Figure 6, the machine tool controller does not measure the actual feedrate at the cutter location point. Therefore, in this paper, the measured feedrate of the point that is closest to the $n^{th}$ cutter location is used as the measured feedrate for the $n^{th}$ cutter location.

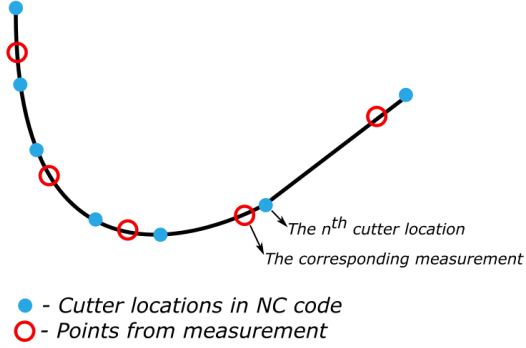

Figure 6. Cutter locations in NC code and cutter locations measured from NC controller.

Table 1 Features for neural networks for feedrate prediction.

| Axis | Feature | Description |
|---|---|---|
|  | $\theta_{n-1}$ | The angle between line segments at the $(n-1)^{th}$ cutter location |
|  | $\theta_n$ | The angle between line segments at the $n^{th}$ cutter location |
|  | $\theta_{n+1}$ | The angle between line segments at the $(n+1)^{th}$ cutter location |
| X | $f_{x,n-1}$ | Nominal feedrate at the $(n-1)^{th}$ cutter location for X axis. |
|  | $f_{x,n}$ | Nominal feedrate at the $n^{th}$ cutter location for X axis. |
|  | $f_{x,n+1}$ | Nominal feedrate at the $(n+1)^{th}$ cutter location for X axis. |
|  | $a_{x,n-1}$ | Nominal acceleration at the $(n-1)^{th}$ cutter location for X axis. |
|  | $a_{x,n}$ | Nominal acceleration at the $n^{th}$ cutter location for X axis. |
|  | $a_{x,n+1}$ | Nominal acceleration at the $(n+1)^{th}$ cutter location for X axis. |
| Y | $f_{y,n-1}$ | Nominal feedrate at the $(n-1)^{th}$ cutter location for Y axis. |
|  | $f_{y,n}$ | Nominal feedrate at the $n^{th}$ cutter location for Y axis. |
|  | $f_{y,n+1}$ | Nominal feedrate at the $(n+1)^{th}$ cutter location for Y axis. |
|  | $a_{y,n-1}$ | Nominal acceleration at the $(n-1)^{th}$ cutter location for Y axis. |
|  | $a_{y,n}$ | Nominal acceleration at the $n^{th}$ cutter location for Y axis. |
|  | $a_{y,n+1}$ | Nominal acceleration at the $(n+1)^{th}$ cutter location for Y axis. |
| Z | $f_{z,n-1}$ | Nominal feedrate at the $(n-1)^{th}$ cutter location for Z axis. |
|  | $f_{z,n}$ | Nominal feedrate at the $n^{th}$ cutter location for Z axis. |
|  | $f_{z,n+1}$ | Nominal feedrate at the $(n+1)^{th}$ cutter location for Z axis. |
|  | $a_{z,n-1}$ | Nominal acceleration at the $(n-1)^{th}$ cutter location for Z axis. |
|  | $a_{z,n}$ | Nominal acceleration at the $n^{th}$ cutter location for Z axis. |
|  | $a_{z,n+1}$ | Nominal acceleration at the $(n+1)^{th}$ cutter location for Z axis. |

*2.3 Training and prediction*

The training process for the neural network model of each axis is shown in Figure 7. The weights of connections in the neural network model are updated during each iteration to decrease the loss function, which is the difference between the predicted feedrate and the measured actual feedrate. In this paper, minimum square error (MSE) was used as the loss function. To apply this process automatically, a backpropagation algorithm [32] was used.

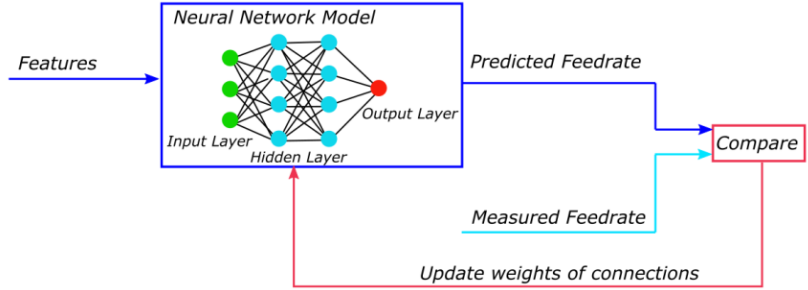

Figure 7. The training process for each axis.

With the trained neural network models, the feedrate of each axis at the $n^{th}$ cutter location can be predicted. The resultant feedrate $f_{n,p}$ can be calculated by equation (9) in which $f_{x,n,p}$, $f_{y,n,p}$ and $f_{z,n,p}$ are the predicted feedrates for the X, Y, and Z axis, respectively. The machining time for an NC code with $N$ numbers of cutter locations can be calculated by equation (10).

$$f_{n,p} = \sqrt{(f_{x,n,p})^2 + (f_{y,n,p})^2 + (f_{z,n,p})^2} \tag{9}$$

$$T = \sum_{n=2}^{N} \frac{\sqrt{(X_n - X_{n-1})^2 + (Y_n - Y_{n-1})^2 + (Z_n - Z_{n-1})^2}}{f_{n,p}} \tag{10}$$

## 3. Experimental Results and Procedure

The machining cycle time prediction method has been experimentally validated using a representative industrial thin-wall structure component.

The research was conducted on a Starrag Scharmann Ecospeed 2538 high-performance machining center controlled by a Siemens 840D. The commercial machining center, designed for machining large aerospace components, has been modified to add a device [27] to measure and output the actual feedrate data. The device connects to the numerical controller (NC) of the machine tool and can read the actual position, speed, and spindle power from the NC of the machine tool. The component for this test and their pocket number are shown in Figure

8. The milling trials were performed under minimum quantity lubrication (MQL) cutting conditions; and the toolpath strategy of the milling operation consisted of a high-performance roughing toolpath for each pocket as shown in Figure 9. The cutting tool diameter was Walter MB266-20.0A3X400B-WJ30UU Ø20x30/R4. The spindle speed was 13300 rpm. The feedrate ranged from 3000 mm/min to 4000 mm/min.

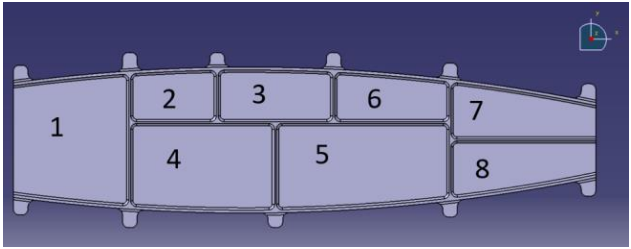

(a)

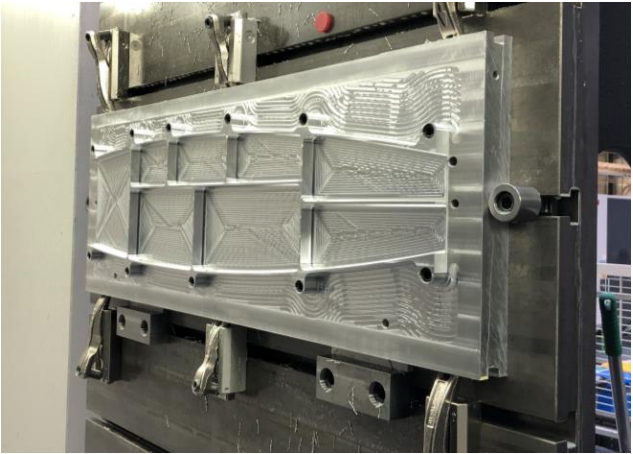

(b)

Figure 8. The Component for machining trials (a) CAD Geometry and pocket number (b) after machining.

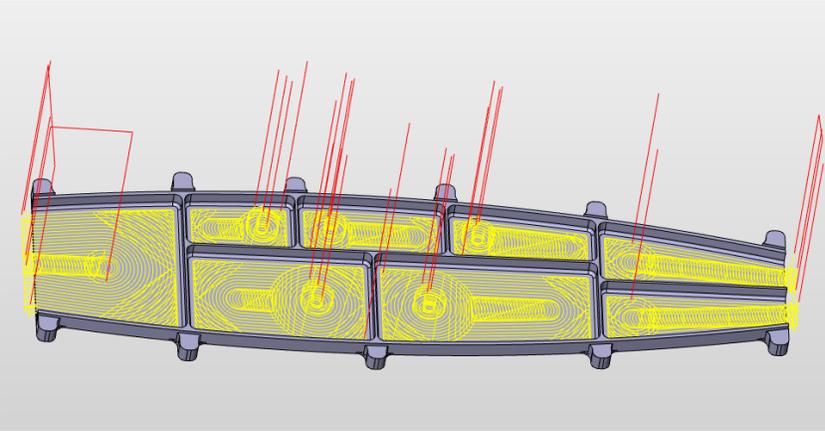

Figure 9. Toolpath for the test component.

# 4. Neural Network Training and Prediction Results

A neural network with two hidden layers was used. The first layer had 5 neurons and the second layer had 10 neurons. It should be noted that a different number of layers and neurons has been tested and the above gave the best result during this experiment. More hidden layers and neurons will make the model capable of learning more complex behaviour of a system. However, more hidden layers will require more data for training and may cause overfitting, which leads to inaccurate prediction. In this experiment, the discussion will be focusing on the application of neural network algorithms in machining cycle time prediction instead of the optimization of the neural network model.

During the machining trials, data was collected for all the pockets. Two neural network models were trained with different datasets to fit the neural network models, as shown in Table 2. The validation dataset was used to prevent overfitting, and the test dataset was used to evaluate the performance of the neural network models during training. To investigate how the dataset could influence the prediction, model #1 only used data from only pocket #1; and model #2 used a combination of data from pocket #1, #5 and #8 as the training dataset. In model #1, pocket #1 was selected because it was the first pocket being machined. In model #2, pocket #5 and pocket #8 were selected because the pocket shape and topology are both different from pocket #1, which could make the model more robust. Both the validation dataset and the test dataset must be different from the training dataset. Therefore, the first half of data from pocket #4 was used as the validation dataset and the rest of the data in pocket #4 was used as the test dataset during the training process.

Table 2 Datasets for the training process.

| Model # | Training dataset | Validation dataset | Test dataset |
|---|---|---|---|
| 1 | Pocket #1 | 50% of pocket #4 | 50% of pocket #4 |
| 2 | Pocket #1, #5, #8 | 50% of pocket #4 | 50% of pocket #4 |

## *4.1 Results from model #1*

Table 3 shows the result of model # 1 with the comparison of the measured and predicted machining cycle time. In general, the prediction of the neural network model was better than the CAD/CAM prediction. The prediction error of the neural network model #1 was less than 6.42% for most of the pockets except for pockets #7 and #8, which presented a prediction error of less than 10% nonetheless. It should be noted that the prediction error for pocket #1 showed a very small error because the neural network model was trained by data from the same pocket #1.

Table 3 Machining time prediction with model #1.

| Pocket # | Measured machining time | CAD/CAM prediction (s) | CAD/CAM prediction error | Neural network model prediction | Neural network model prediction |
|---|---|---|---|---|---|

|   | (s) |     | (%)   | (s) | error (%) |
|---|-----|-----|-------|-----|-----------|
| 1 | 308 | 273 | 11.36 | 310 | 0.65      |
| 2 | 78  | 63  | 19.23 | 73  | 6.41      |
| 3 | 107 | 90  | 15.89 | 101 | 5.61      |
| 4 | 205 | 187 | 8.78  | 202 | 1.46      |
| 5 | 214 | 191 | 10.75 | 205 | 4.21      |
| 6 | 109 | 92  | 15.60 | 102 | 6.42      |
| 7 | 160 | 129 | 19.38 | 148 | 7.50      |
| 8 | 153 | 127 | 16.99 | 139 | 9.15      |

Figure 10(a) shows the results of feedrate prediction for pocket #1. Though the command feedrate was relatively constant for G0 and G01 respectively, the actual feedrate was fluctuating and the neural network model was able to capture that feature. For example, as shown in Figure 10(b) and Figure 10(c), for the G0 command in the NC code (30000 mm/min), the actual feedrate cannot reach that speed and the neural network model learned that behavior successfully. Also, as the details of the feedrate prediction shown in Figure 10(b) and Figure 10(c), the neural network learned to predict the actual feedrate for G01 commands as well.

Figure 11(a) shows the results of feedrate prediction for pocket #3. From the details shown in Figure 11(b) and Figure 11(c), it shows that the neural network model was able to predict the change of feedrate for the data that is not from the training or the validation set.

At some cutter locations, it can be seen that there is still a difference between the predicted feedrate and the measured feedrate. This may be because of the low sampling rate of the measurement system which was restricted to around 5 Hz, which introduced the inaccuracy. As shown in Figure 12, the number of measured points (901 points) was much less than that of the commanded points (5383 points) in the NC code. Therefore, the feedrate at measured points might be slightly different from those at the commanded points, which introduced error into the model.

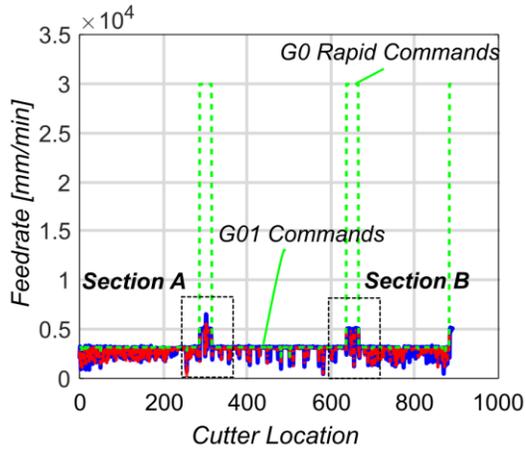
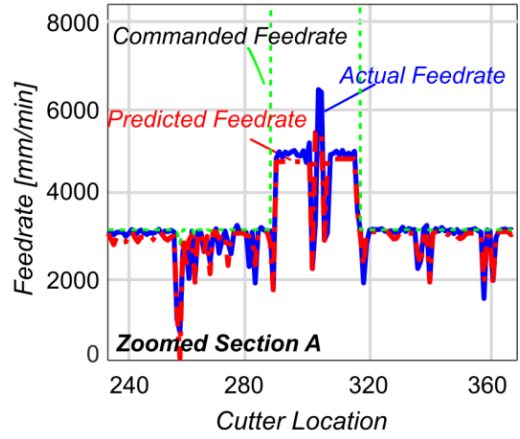
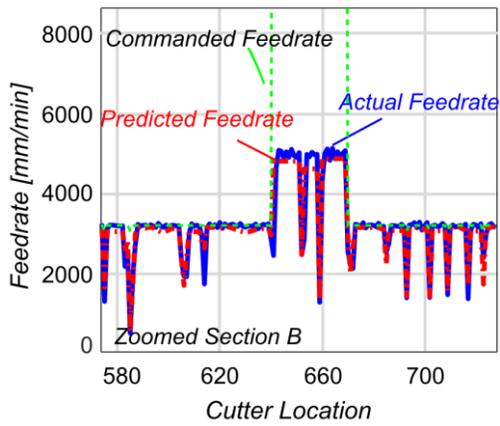

Figure 10. Resultant feedrate results with model #1 for (a) pocket #1 (b) zoomed section A (c) zoomed section B.

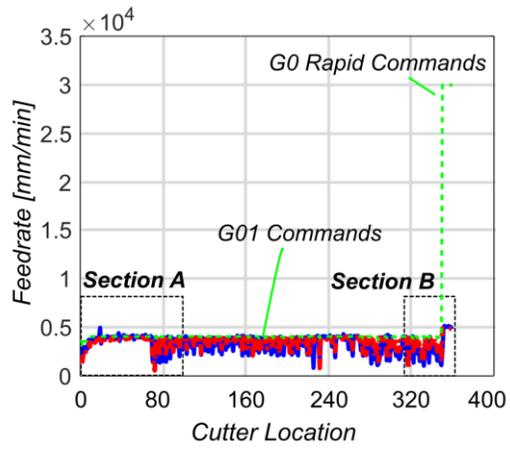
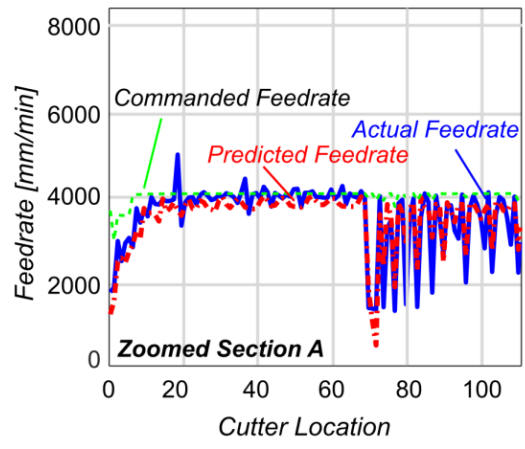

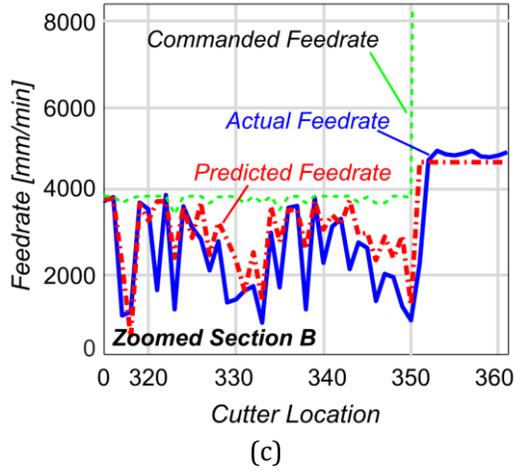
(c)
Figure 11. Resultant feedrate results with model #1 for (a) pocket #3 (b) zoomed section A (c) zoomed section B.

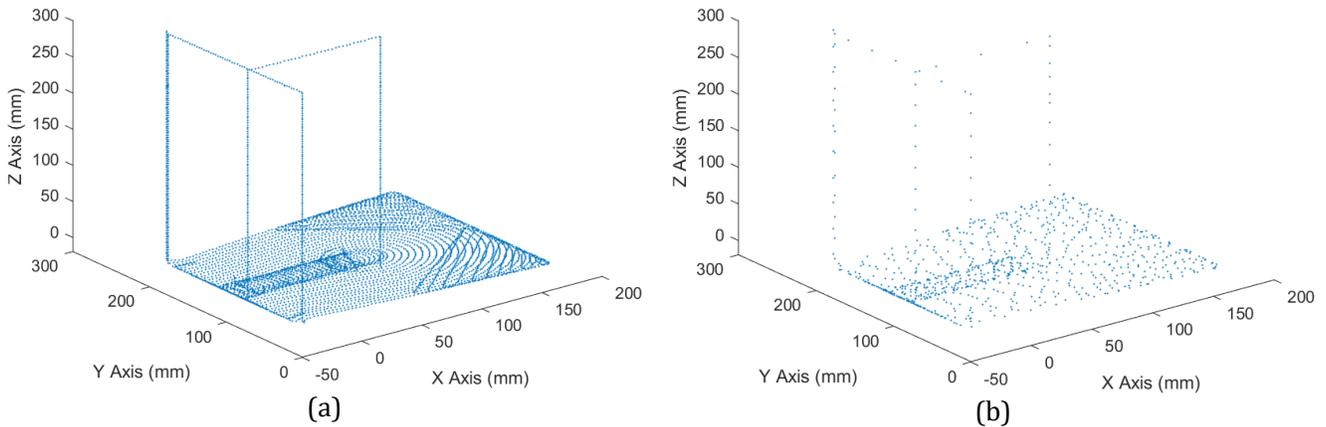
Figure 12. Toolpath for pocket #1 (a) Cutter locations in CAM (b) Cutter locations measured from the controller.

Figure 13(a) shows the prediction results for pocket #8 with the neural network model #1 which was trained with only pocket #1 data. As the details of the feedrate prediction shown in Figure 13(a), for most of the G01 command, the neural network did a relatively accurate prediction of the feedrate. However, for the G0 command at the beginning of the tool path (Figure 13(b)), the neural network was not able to predict the feedrate correctly because it was difficult for a data-driven model to accurately predict the feedrate when it met a new circumstance. Figure 14(a) shows that the G0 command in X-axis for pocket #1 is negative, and therefore, the model only learned how to predict under this circumstance. However, in pocket #8 data (Figure 14(b)), the G0 command in X-axis was commanded negative first and then positive, which was a new case for model #1. Therefore, model #1 cannot predict it correctly. Then in section 4.2, model #2 was used and compared with model #1.

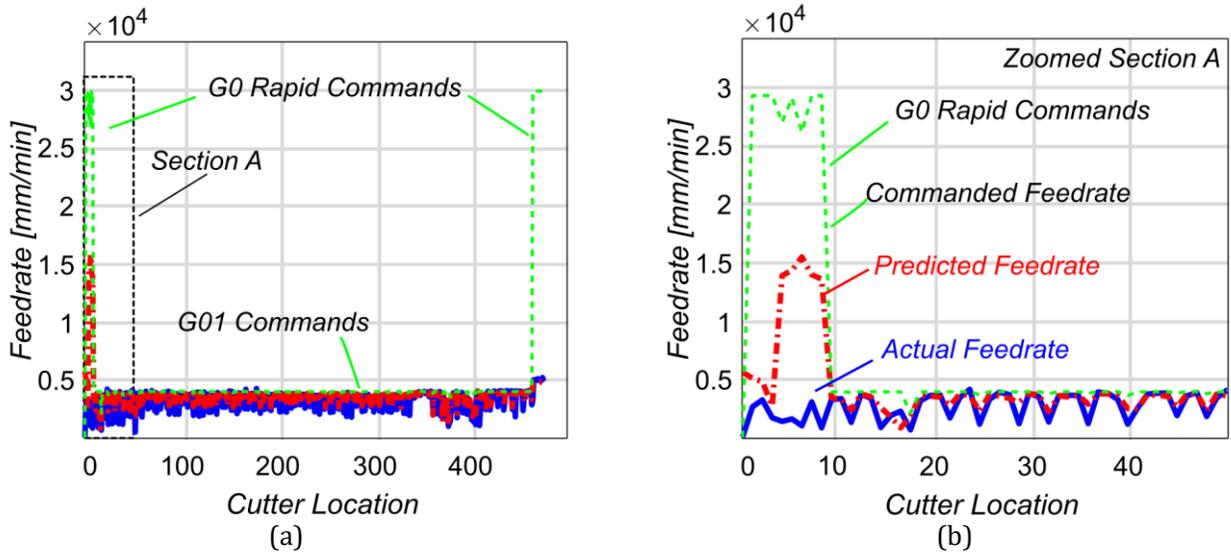

Figure 13. Resultant feedrate results with model #1 (a) for pocket #8 (b) zoomed section A.

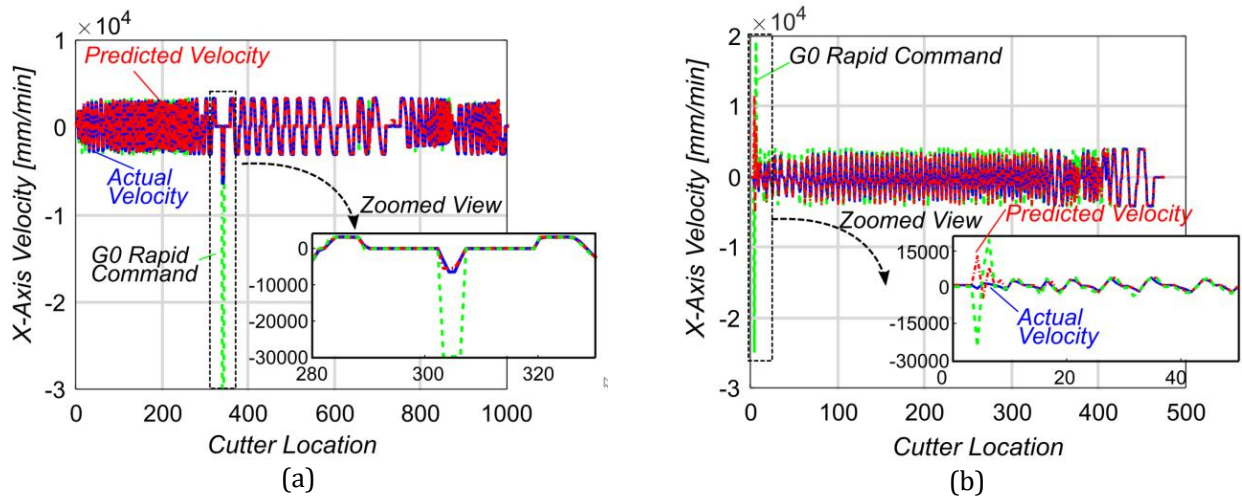

Figure 14. Feedrate in X axis with model #1 (a) Pocket #1 (b) Pocket #8.

### *4.2 Results from model #2*

As stated in Table 2, model #2 was trained with more data. Data from pocket #1, pocket #5 and pocket #8 was used to train model #2. Figure 15 shows the predicted feedrate for pocket #8 by model #2. Compared to Figure 13(b), Figure 15 (b) shows the prediction results from model #2 for the G0 commands of pocket #8 were improved. As shown in Figure 16, model #2 also improved the prediction for the G0 command for pocket #7 which has a similar shape to pocket #8.

Moreover, the overall cycle time prediction accuracy was also improved. Table 4 shows the machining cycle time prediction when the neural network model # 2 was trained with data from pocket #1, pocket #5 and pocket #8. The prediction error was no more than 3%. Though the error for pockets #1 and #4 was slightly increased, the error for other pockets was improved, which means the model was more generalized. This means the neural network model can keep improving itself by learning more data during the manufacturing process.

It should be noted that by learning more data, the model was generalized and avoided overfitting on one pocket. Therefore, though the model was trained on #1, #5 and #8, they may not have the minimum error for the prediction, which is different from the case for model #1. It should also be noted that it is a coincidence that pocket #3 has a 0% error.

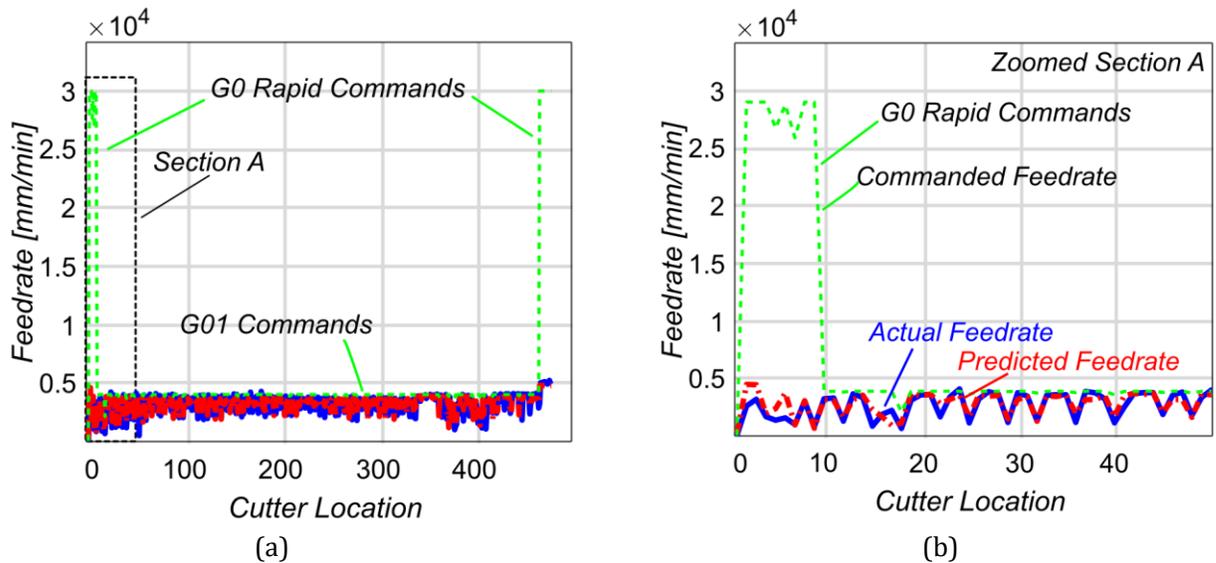

Figure 15. Resultant feedrate results for pocket #8 (a) by model #2 (b) zoomed section A.

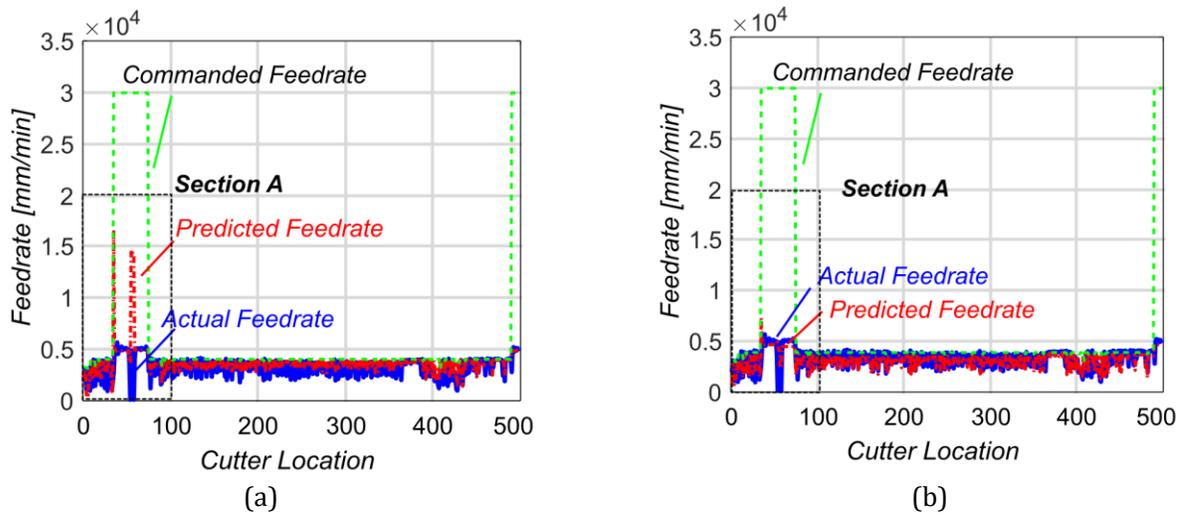

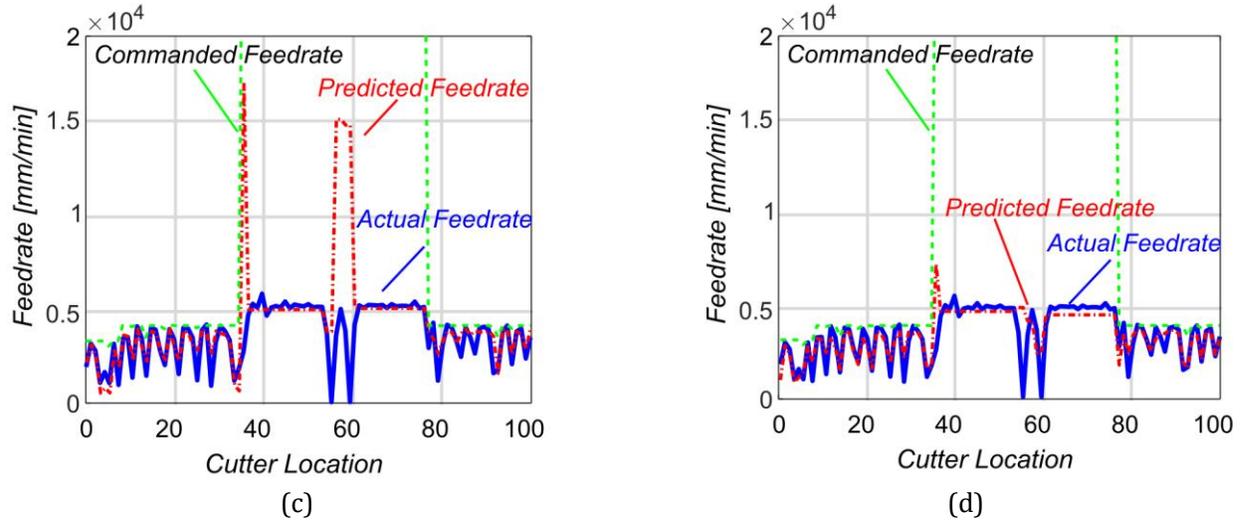

Figure 16. Resultant feedrate results for pocket #7 (a) by model #1 (b) by model #2 (c) section A with model #1 (d) section A with model #2.

Table 4 Machining time prediction with model #2.

| Pocket # | Measured machining time (s) | Neural network model prediction (s) | Neural network model prediction error (%) |
|---|---|---|---|
| 1 | 308 | 314 | 1.95 |
| 2 | 78 | 77 | 1.28 |
| 3 | 107 | 107 | 0 |
| 4 | 205 | 211 | 2.93 |
| 5 | 214 | 216 | 0.93 |
| 6 | 109 | 106 | 2.75 |
| 7 | 160 | 158 | 1.25 |
| 8 | 153 | 149 | 2.61 |

## 5. Conclusion and Future Work

This paper has presented a novel feedrate and machining cycle time prediction method. The main findings are as follows:

1. The research presented a data-driven method to model the behavior of a complex machine tool system. Experiment results showed that this method estimated the machining cycle time with more than 90% accuracy, which is comparable to other analytical methods.

2. The overall accuracy of the feedrate and cycle time prediction by the neural network model can be improved by training with more data. For a new circumstance that the model has never met, it may not predict the result properly. However, after training and updating the model with the data from the new circumstance, the model can learn to predict a better result. This means the model can keep improving itself in time by learning from more data.
3. In this paper, the features for the neural network model were selected as per commonly used in the literature for physics-based models. In the future, automatic feature selection and generation methods will be explored.

This method showed that neural network models could learn the behavior of a complex machine tool system, which could be served as a model for machining time prediction. Further research will extend the applications to 5-axis toolpaths and higher accurate prediction with a high sampling rate system.

## Acknowledgment

The authors would like to acknowledge the funding provided by High Value Manufacturing Catapult which supported this project.